# The Vehicle Trajectory Prediction Based on ResNet and EfficientNet Model

Ruyi Qu[1]
University of Toronto,
Toronto, Canada,
ruyi.qu@mail.utoronto.ca

ShuKai Huang[1]
University of Brimingham,
Birmingham, United Kingdom
sxh1245@alumni.bham.ac.uk,

Jiexuan Zhou[2],
Guangzhou Huali College,
Guangzhou, China,
GlacierMelt@163.com,

ChenXi Fan[3],
Shanghai Ocean University,
Shanghai, China,
646645399@qq.com,

ZhiYuan Yan[*],
Zhongshang Inc,
shenzhen，China,
810824494@qq.com,

*Abstract*—At present, a major challenge for the application of automatic driving technology is the accurate prediction of vehicle trajectory. With the vigorous development of computer technology and the emergence of convolution depth neural network, the accuracy of prediction results has been improved. But, the depth, width of the network and image resolution are still important reasons that restrict the accuracy of the model and the prediction results. The main innovation of this paper is the combination of RESNET network and efficient net network, which not only greatly increases the network depth, but also comprehensively changes the choice of network width and image resolution, so as to make the model performance better, but also save computing resources as much as possible. The experimental results also show that our proposed model obtains the optimal prediction results. Specifically, the loss value of our method is separately 4 less and 2.1 less than that of resnet and efficientnet method.
.

## I. Introduction

As an important means of transportation for mankind, vehicles have a history of one hundred years. In the traditional driving environment, the driver makes analysis and judgment by observing the road conditions and pedestrians around the vehicle, so as to control the vehicle driving, which has great uncertainty [1]. At the same time, in recent years, with the continuous improvement of people's living standards, the number of vehicles is also increasing, which leads to the traffic situation becoming more complex. For example, forced lane change and traffic congestion of vehicles may cause traffic accidents or traffic jams, and it is difficult for drivers to make accurate response and judgment in time [2]. The emergence of automatic driving technology can make the vehicle automatically perceive the surrounding driving environment. When encountering obstacles such as pedestrians and bicycles, the vehicle can make rapid judgment according to the predetermined rules of the control system, plan a collision free path or the path with the lowest risk, and control the vehicle movement until it reaches the set destination, which not only liberates the driver, Moreover, it improves the driving safety and traffic order to a great extent [3,4].

At present, many countries and companies are actively involved in the research of automatic driving technology and have achieved certain results [5], such as Google, Tesla, Baidu, etc. However, there are still great engineering challenges to be solved before realizing automatic driving in all aspects. One of the challenges is to establish an accurate vehicle trajectory prediction model [6], which can quickly and accurately judge and plan the driving route when there are obstacles such as pedestrians, bicycles or other vehicles around the vehicle, so as to avoid traffic accidents caused by algorithm or model errors. Therefore, the related technologies or algorithms of automatic driving have become the research focus and hotspot in the automotive field at this stage. Many mathematical prediction models have been applied to the prediction of vehicle trajectory, mainly including linear regression prediction model [7], neural network model [8], Markov model [9] and so on. However, the main disadvantage of the above traditional algorithms is that it is difficult to build a very complex nonlinear model, which will lead to low prediction efficiency and large error, which is difficult to be applied to practice.

With the vigorous development of computer technology and the rapid increase of data, many deep learning algorithms are widely used to solve complex problems in all walks of life. Tan *et al.* [10] proposed a model coupling the improved limit learning machine (ELM) and deep neural network to predict the vehicle trajectory. The experimental results show that the algorithm has higher prediction accuracy and efficiency than the traditional algorithm. Zhang *et al.*[2] used LSTM method to extract the characteristic parameters during vehicle driving, predicted the future driving trajectory of the vehicle, and obtained more accurate results. This paper presents a vehicle trajectory prediction network based on the combination of EfficientNet and ResNet algorithms. Compared with traditional neural networks, ResNet algorithm can greatly improve the accuracy of prediction by increasing the depth of neural networks, such as training hundreds or even thousands of neural networks, and alleviate the problem of gradient disappearance caused by the increase of depth [11].

The EfficientNet model can comprehensively adjust the size of the input image, the depth and width of the neural network, so as to achieve a higher accuracy than other Neural Network algorithm, fully save computing resources and improve the efficiency of the model [12]. We coupled the above two algorithms and obtained the best vehicle trajectory prediction performance. Specifically, the minimum loss function of the proposed method after predicting the vehicle trajectory is 21.82, 4.00 less than RESNET algorithm and 2.16 less than EfficientNet algorithm, which shows that the predicted value of our algorithm is closer to the real value and has better robustness.

In general, the key contributions of this paper are as follows:

● We combine ResNet and efficient net algorithm for the first time, and obtain the best vehicle trajectory prediction performance compared with other algorithms.

● Through the combination of ResNet and Efficient Net networks, we not only greatly deepen the depth of the network, but also effectively reduce the computing resources and simulation time required by the experimental operation.

The structure of the rest of this paper is as follows: in the second section, the related work and research status of vehicle trajectory prediction are introduced; The third section describes in detail the algorithm model and deep neural network architecture used in this paper; The fourth section introduces the data set used in the experiment and shows the performance of our algorithm and other neural network algorithms in vehicle trajectory prediction. Finally, in the fifth part, the research work of this paper is summarized and the future work is prospected.

## II. RELATED WORK

### A. Vehicle trajectory

In recent years, algorithms such as machine learning and deep neural network have been widely used in the prediction of vehicle trajectory. Traditional machine learning algorithms, such as Bayesian network, Markov chain, Gaussian mixture and mining frequent trajectory patterns, have been applied to vehicle trajectory prediction. Jureee et al. [13] predicted the uncertain trajectory of vehicles by the universality of Gaussian model and the advantage that Bayesian network can overcome the local optimal solution of expectation maximization algorithm. Qiao et al. [14] Studied the Gaussian mixture model trajectory prediction method, and the results show that this method has high accuracy. Li et al. [15] Combined Bayesian network and Markov model to predict the vehicle trajectory, which show that this method performs poorly in the case of discrete trajectory. The emergence of artificial neural network can enable people to establish more complex nonlinear system models and predict more complex problems. Ding et al. [16] used BP neural network to predict the lane change trajectory of vehicles based on the data of passing vehicles. The test results show that BP neural network can accurately predict the lane change behavior of vehicle drivers.

Relevant research shows that as long as the number of hidden layers in a neural network is enough, the model can approach a nonlinear function with arbitrary accuracy, that is, the predicted value can approach the real value infinitely. When there are too many neurons in the hidden layer, it would lead to the questions of disappearance or explosion of gradient, network degradation and so on. The emergence of RESNET network solves the problem of network degradation, and expands the number of layers of neural network from more than ten to hundreds or even thousands, which greatly improves the accuracy of deep neural network model prediction [11]. The EfficientNet network comprehensively considers the interaction among the three dimensions of network depth, network width and input image resolution, and gives the best combination of the three [12], which greatly improves the accuracy of prediction under the condition of saving computing resources as much as possible. Therefore, in this paper, we combine the two most advanced depth neural network models for the first time, and get the most accurate prediction results.

## III. Methodology

In this section, we systematically introduce the network algorithm model used in this paper, including ResNet and EfficientNet networks.

Compared with the traditional depth neural network, ResNet network introduces a new network structure called "residual block" in order to alleviate the gradient disappearance problem [17], as shown in Figure 1(b). The main reason for network degradation is that with the increase of network depth, some irreversible information loss would be caused every time from input to output. Therefore, to keep the network from degradation, the fundamental lies in making the output h equal to the input x when the network level is deep. In traditional neural networks, it is difficult to fit the potential identity mapping function $H(x) = x$; However, for the residual module, the input and output are directly added together, $H = F + X$ can be obtained. At this time, the problem of identity mapping function $H = x$ is transformed into the problem of residual function. Moreover, fitting the residual function is much easier than fitting the identity mapping function. Therefore, ResNet network can effectively prevent network degradation.

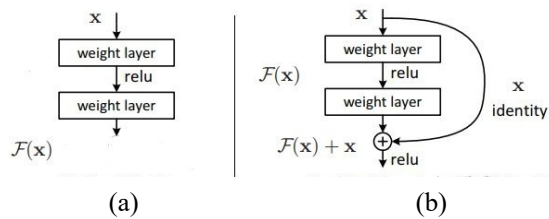

Figure 1. The original (a) and residual block (b)

The function $F(x, \{W_i\})$ represents the residual mapping needed to learn.

And then the EfficientNet network is introduced. Some scholars have shown that simply increasing a dimension in the depth neural network has certain limitations in improving the accuracy of prediction results. The dimensions of model expansion are not completely independent. For example, for deeper networks, wider networks and greater image resolution should be used, which means that it is necessary to balance the expansion dimensions rather than expand in a single dimension. Therefore, the EfficientNet method appears. It is a compound expansion method, which controls the depth (*d*) and

width ($w$) of the network and the resolution ($r$) of the image through parameters (α, β, γ), shown as:
$$d = \alpha^\emptyset$$
$$w = \beta^\emptyset$$
$$r = \gamma^\emptyset$$
$$\alpha \cdot \beta^2 \cdot \gamma^2 \approx 2$$
where $\alpha \geq 1, \beta \geq 1, \gamma \geq 1$. In the initial stage, $\emptyset$ is fixed to 1, and then the optimal parameters $\alpha$, $\beta$ and $\gamma$ are obtained by grid search method, which means that the minimum optimal basic model is obtained. Then, fix the values of $\alpha$, $\beta$ and $\gamma$ and continuously increase the value of $\emptyset$, that is, it is equivalent to expanding the three dimensions of the model at the same time, which will improve the performance and increase the resource consumption.

In this paper, the above two network models are combined to predict the vehicle trajectory. The mixing mode of the two models is shown in Figure 2:

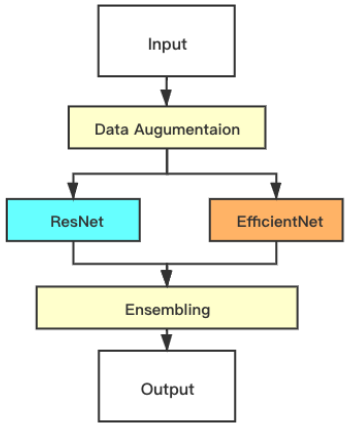

Figure 2. The mixed model

## IV. Experiments

### A. Experimental Data and Settings

In this part, we first introduce the data set used in the experiment. The data used in this paper is part of the large amount of data released by LYFT on the KAGGLE competition platform. These data are used to establish the motion trajectory prediction model for autonomous vehicles, which contains a set of scenes (driving episodes acquired from a given vehicle), frames (snapshots in time of the pose of the vehicle), agents (a generic entity captured by the vehicle's sensors), agents_mask (a mask that (for train and validation) masks out objects that aren't useful for training), and traffic light information.

In this paper, we mainly extract the vehicle position information in the picture and convert the vehicle position into two-dimensional coordinates, as shown in Figure 3. By extracting the historical vehicle trajectory data, we can get the change of vehicle position coordinates with time and the change law of vehicle trajectory coordinates when there are obstacles around the vehicle, such as pedestrians, bicycles or other vehicles.

Then, we use the convolution depth neural network model, combined with RESNET and efficient net network structure to optimize the model, train a large amount of data, and get the prediction results of vehicle trajectory.

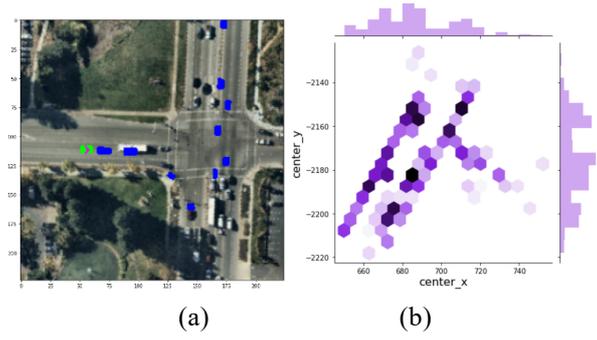

Figure 3. Vehicle location information

### B. Experimental Results and Analysis

In this section, we first give the vehicle trajectory predicted by the proposed algorithm. In the experiment, the selected model parameters are as follows: the learning rate is 1e-5, the batch size is 16, and the Radam optimizer is preferred, and the results are shown in Figure 4.

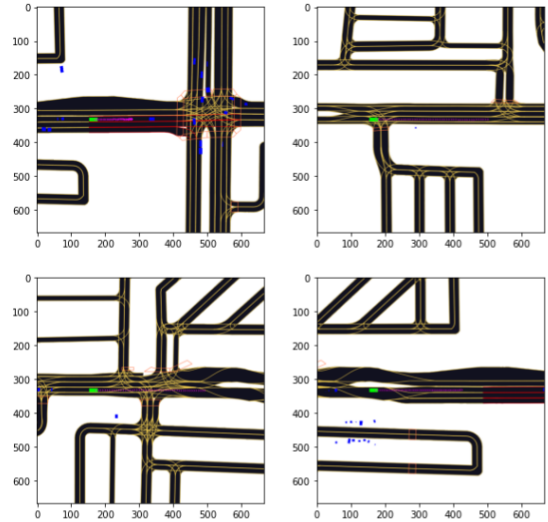

Figure 4. The results of prediction

In order to comprehensively measure the performance of our proposed algorithm, we select the value of loss function as the index, which represents the gap between the predicted and the real values. Generally speaking, the smaller the value of this index, the better the algorithm model and the more robust the algorithm is. And the loss function used here can be written as:

$$L = -log p(x_{1,\dots,T}, y_{1,\dots,T} | c^{1,\dots,K}, \overline{x}_{1,\dots,T}^{1,\dots,K}, \overline{y}_{1,\dots,T}^{1,\dots,K})$$
$$= -log \sum_k e^{\log(c^k) - \frac{1}{2}\Sigma_t(\overline{x}_t^k - x_t)^2 + (\overline{y}_t^k - y_t)^2}$$

where ($x_{1,\dots,T}, y_{1,\dots,T}$) represent the ground truth positions of a sample trajectory, and ($\overline{x}_{1,\dots,T}^{1,\dots,K}, \overline{y}_{1,\dots,T}^{1,\dots,K}$) stand for the $k$ hypotheses predicted, represented by means.

The loss function values of several algorithms are given in Table 1. Obviously, our algorithm has the smallest index of 21.82, which is 15.49% smaller than RESNET method and 8.93% smaller than EfficienNet algorithm, which shows that our algorithm has the best performance in predicting vehicle trajectory.

| Models | Loss |
|---|---|
| ResNet | 25.82 |
| EfficientNet | 23.96 |
| Proposed | 21.82 |

Table 1. Comparison between loss functions of different algorithms

## V. Conclusions

Based on RESNET grid and efficient net grid model, a new convolution depth learning model for vehicle trajectory detection is established in this paper. The model has a deep network level, and has the optimal network width and image resolution. Experimental results show that our model has higher prediction accuracy and better robustness than other deep learning algorithms.

In the future, we plan to improve the ability of the algorithm to continuously reduce the prediction error of vehicle trajectory, so as to provide some guidance for the application of automatic driving technology and future scientific research.